\documentclass[runningheads]{llncs}

\usepackage[T1]{fontenc}

\usepackage{graphicx} 
\usepackage{booktabs} 
\usepackage{array}    
\usepackage{ragged2e} 
\usepackage{rotating} 
\usepackage{url} 
\usepackage{hyperref} 
\usepackage{color}

\begin{document}
%
\title{The Emerging Paradigm of\texorpdfstring{\\}{ }Geospatial Foundation Models:\texorpdfstring{\\}{ }From Pre-Training to Agentic Reasoning}
\titlerunning{The Emerging Paradigm of Geospatial Foundation Models}
\author{Shelley Cazares\orcidID{0000-0003-0685-1626}}
\authorrunning{S. Cazares}
\institute{Google Public Sector, Washington DC 20001, USA\\
\email{shelleycazares@google.com}}
\maketitle
%
\begin{abstract}
The analysis of satellite and aerial imagery has entered a new era with the advent of foundation models. This paper describes the concept of Geospatial Foundation Models (GeoFMs), which are artificial intelligence/machine learning (AI/ML) models pre-trained on massive geospatial datasets through varied methodologies. We first articulate the core paradigm shift that GeoFMs enable: a separation of duties, where large-scale model providers perform the computationally intensive pre-training, allowing domain experts to rapidly fine-tune or prompt these models for specific, mission-critical tasks. This approach democratizes access to state-of-the-art AI/ML while maintaining the security and confidentiality of the downstream task. We then explore the novel capabilities unlocked by different types of GeoFMs, distinguishing between the fine-tunable vision models produced by self-supervised techniques like masked auto-encoding, and the vision-language models produced by contrastive learning which enable zero-shot tasks like open-vocabulary image analysis. Next, we discuss the practical considerations for operationalizing GeoFMs, from performance-cost analysis to the broader MLOps ecosystem. To that end, we introduce a taxonomy of model adaptation strategies and propose a framework for domain experts to select the most cost-effective adaptation approach for their particular mission set. Finally, we present a forward-looking vision of Agentic Geospatial Reasoning, where Large Language Models act as intelligent orchestrators, leveraging GeoFMs as tools to answer high-level user queries in natural language and automate complex analytical workflows, moving the field from perception to cognition.

\end{abstract}

\section{Introduction}
The increasing volume and variety of Earth observation data present both a monumental opportunity and a significant challenge. While satellite and aerial imagery provide a rich source of information for applications ranging from disaster response to environmental monitoring, the ability to analyze this data at scale and speed remains a critical bottleneck. Traditional supervised machine learning approaches, while powerful, often require large, meticulously labeled datasets for each specific task. This requirement represents a significant bottleneck, as the cost and time associated with data curation and labeling are often untenable for applications demanding rapid deployment.

A compelling example of this challenge is rapid damage assessment following a natural disaster. In the aftermath of an event like a category 5 hurricane, response teams need immediate, actionable intelligence to prioritize resource allocation. The limitations of the traditional supervised learning approach have been well documented \cite{skai}. The data curation and labeling process for such an endeavor could take days or even weeks, an untenable timeline in a crisis. To address this bottleneck, research has demonstrated that semi-supervised learning can mitigate this challenge by achieving strong performance with only a fraction of the labeled data for a particular geographical area \cite{skai}.

While semi-supervised learning reduces the in-crisis labeling burden, this paper explores an emerging paradigm: the Geospatial Foundation Model (GeoFM), which takes this principle a step further. Instead of training a model on an unlabeled dataset gathered \textit{reactively} from a single disaster area, a GeoFM is pre-trained \textit{proactively} on a global scale for general use. A GeoFM is an artificial intelligence/machine learning (AI/ML) model pre-trained on a massive, varied corpus of geospatial data collected from all over the globe, which can then be adapted (fine-tuned or prompted) to perform a wide range of downstream geospatial tasks. This pre-trained global knowledge means that the GeoFM is \textit{not} starting from scratch when a new crisis occurs, allowing for even greater data efficiency during adaptation.

This proactive approach enables the most significant contribution of the GeoFM paradigm: A separation of duties. Computationally intensive, general-purpose pre-training is performed by large organizations with sufficient resources, using unlabeled or weakly-labeled data that does not require access to downstream, task-specific information. Domain experts can then perform task-specific adaptation using their own small set of proprietary labels, ensuring the security and confidentiality of their sensitive data and use cases. This paradigm democratizes access to cutting-edge AI/ML for the applied geospatial imagery community while maintaining the security and confidentiality of the downstream task.

This paper focuses on overhead imagery-based GeoFMs, those which are used to process satellite and airborne imagery with a downward-facing look angle. We will first define two dominant pre-training strategies for GeoFMs: self-supervised and contrastive learning. We will then survey the new capabilities unlocked by these distinct pre-training strategies. Next, we will cover the practical aspects of operationalizing these models. Finally, we will conclude with a vision for the future: Agentic Geospatial Reasoning, where multimodal Large Language Models (LLMs) act as intelligent orchestrators, leveraging GeoFMs as tools to automate complex analytical workflows directly from natural language prompts.

\section{The Geospatial Foundation Model Paradigm}
Image-based GeoFMs typically leverage the vision transformer architecture \cite{16x16}, whose self-attention mechanism is exceptionally well-suited to modeling the long-range spatial dependencies inherent to satellite and aerial imagery. These models are pre-trained on massive datasets curated to be varied across geographies, seasons, and times of day, aiming to build a comprehensive representation of the earth's surface. The core innovation of the foundation model paradigm lies in its two-stage lifecycle: a general-purpose pre-training stage followed by a task-specific adaptation stage. The pre-training strategy fundamentally determines the model's capabilities. Two dominant approaches have emerged: self-supervised pre-training for vision models and weakly-labeled contrastive pre-training for vision-language models.

\subsection{Pre-Training Strategy 1: Self-Supervised Learning for Vision Models}
\begin{figure}[htbp]
\centering
\includegraphics[width=\textwidth]{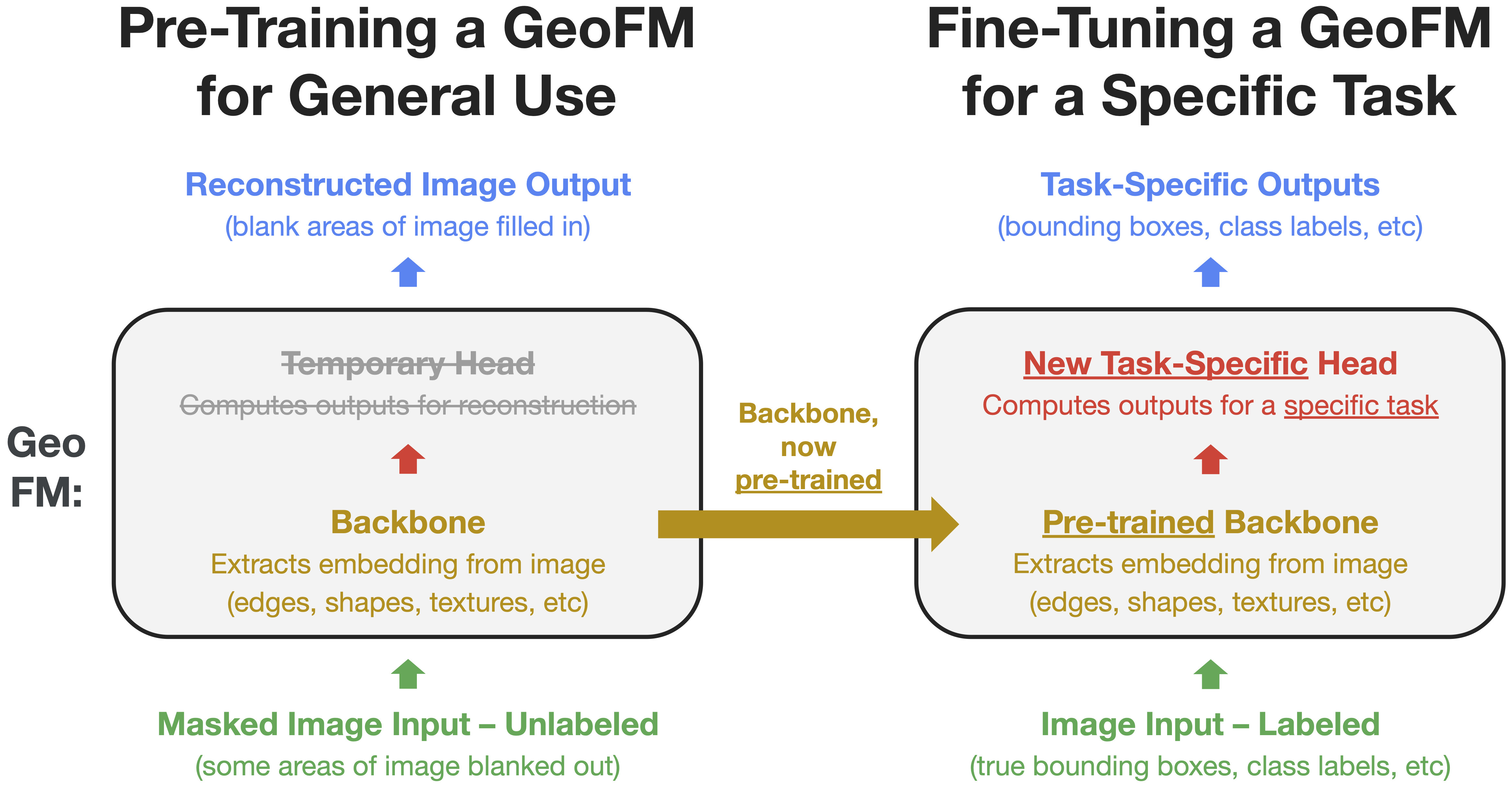}
\caption{The two-stage life-cycle of a vision geospatial foundation model. (Left:) The self-supervised pre-training phase, where a model backbone learns to extract powerful embeddings by reconstructing masked patches with the help of a temporary head. (Right:) The supervised fine-tuning phase, where the pre-trained backbone is used, its temporary reconstruction head discarded, and a new, task-specific head is trained on a small, labeled dataset.} 
\label{pretrain_vs_finetune}
\end{figure}

The first approach focuses on learning robust visual features from images alone, often with self-supervised learning (SSL) methods like Masked Auto-Encoding (MAE) \cite{mae}. The model consists of a backbone (encoder) and a temporary head (decoder). It is trained end-to-end to perform a pretext task: Reconstruct randomly masked patches of an image from the remaining visible patches (Fig.~\ref{pretrain_vs_finetune}, left). This pre-training process, driven by pixel-level reconstruction loss and requiring no human-provided labels, compels the backbone to compress the visible information in the  image into a powerful, general-purpose feature representation (embedding) that the head can then use to complete the reconstruction.

While the reconstruction task is the mechanism for pre-training, the resulting backbone is the developer's goal. Once pre-training is complete, the temporary reconstruction head is discarded, leaving a powerful, pre-trained backbone ready for downstream application. A user can then append a new, task-specific head to the pre-trained backbone and fine-tune it using supervised learning on a small, labeled dataset (Fig.~\ref{pretrain_vs_finetune}, right). This efficient workflow is the central promise of the foundation model paradigm.

\subsection{Pre-Training Strategy 2: Contrastive Learning for Vision-Language Models}
The second dominant pre-training approach focuses on aligning visual features with human language. This is accomplished through contrastive learning techniques (e.g., CLIP \cite{clip}, SigLIP \cite{siglip}) on a massive, weakly-supervised dataset of (image, text caption) pairs. During pre-training, the model learns to map both an image and its corresponding text description (its weak label) to a shared embedding space. The training objective is to maximize the similarity (e.g., cosine similarity) of the embeddings for a correct pair, while minimizing the similarity for incorrect pairs within a batch.

This pre-training process creates a model that understands the semantic relationship between visual content and text. It is this vision-language alignment that enables powerful zero-shot capabilities, such as open-vocabulary image analysis, where the model can generalize to concepts not seen during pre-training. Consequently, direct prompting is the primary mode of application of these contrastively-trained GeoFMs.

\subsection{The Overarching Benefit: A Separation of Duties}
The separation of duties remains a key benefit of the GeoFM paradigm, regardless of the pre-training strategy. Industrial research labs and other large-scale model providers can absorb the immense computational cost of pre-training, enabling the broader community to achieve state-of-the-art performance. Domain experts can either fine-tune the vision models, effectively transferring learned embeddings to new tasks, or directly prompt the vision-language models for powerful zero-shot analysis.

Specifically, the separation of duties offers two significant advantages to domain experts:
\begin{itemize}
    \item \textbf{Security and confidentiality:} Because pre-training can be accomplished with unlabeled or weakly-labeled data, the large organizations who pre-train the GeoFMs do not require access to the domain experts' proprietary labels or use cases, ensuring the security and confidentiality of their sensitive missions.
    \item \textbf{Reduction in labeling resources:} While fine-tuning a GeoFM still requires a labeled dataset, the required dataset size is typically much smaller than that needed to train a bespoke model from scratch. This drastically reduces the cost and time associated with data curation and labeling, making state-of-the-art performance more accessible.
\end{itemize}

\subsection{A Survey of Existing Geospatial Foundation Models}
GeoFMs are a rapidly evolving field, with a vibrant ecosystem including prominent models from both industrial and academic research groups. Notable early models include SatMAE from Stanford \cite{satmae-paper,satmae-code}, which pioneered the adaptation of the MAE concept specifically for geospatial data. It introduced novel temporal and spectral positional encodings to effectively pre-train a transformer backbone on 10m - 60m resolution, multi-spectral, time-series imagery from Sentinel-2 satellites.

Subsequent research has focused on improving the MAE framework's ability to handle the scale variance of remote sensing data, where the same object type can appear at vastly different resolutions given the imagery source. ScaleMAE \cite{scalemae-paper,scalemae-code} addresses this challenge by introducing a scale-aware position encoding based on the image's Ground Sample Distance (GSD), as well as a novel decoder to reconstruct images at multiple scales to explicitly learn multi-scale features. More recently, Cross-Scale MAE \cite{crossscalemae-paper,crossscalemae-code} proposed an alternative approach by using scale augmentation during pre-training and enforcing cross-scale consistency with both contrastive and generative (reconstruction) losses, removing the dependency on known GSD metadata.

Another recent approach has been to build more comprehensive, geospatial-native models. Prithvi-EO-2.0, developed by IBM and NASA \cite{prithvi-paper,prithvi-weights,prithvi-code}, is a MAE-based model pre-trained on a global dataset of 30m resolution Harmonized Landsat and Sentinel-2 (HLS) imagery, explicitly incorporating time and location data into its embeddings to enhance performance on Earth observation tasks. Similarly, Clay \cite{clay} is a MAE-based model designed to accept a wide variety of sensor data with different resolutions and spectral bands, along with time and location data.

Yet another approach has been to leverage massive, non-geospatial datasets. Meta's DINOv3 \cite{dino-paper,dino-code} is a generalist vision model pre-trained on billions of web images using self-distillation (a discriminative SSL approach). It has demonstrated remarkable transfer learning performance on dense geospatial tasks like segmentation. Its success proves that features learned from natural images can, at a sufficient scale, generalize effectively to overhead perspectives. The authors also demonstrated the versatility of their pre-training recipe by training a separate DINOv3 model from scratch on high-resolution (0.6m) Maxar satellite imagery, outperforming many domain-specific models.

Finally, some models are designed to produce a new kind of data product: A universal, analysis-ready "embedding field". Google DeepMind's AlphaEarth Foundations \cite{alphaearth-paper,alphaearth-gee} exemplifies this. It uses a multi-loss training strategy, combining reconstruction, self-distillation, and text-contrastive objectives, to process multi-modal inputs (optical, radar, LiDAR, text, climate data) into a time-continuous representation. The final product is \textit{not} a model to be fine-tuned or prompted with new imagery, but a global 10m resolution dataset of embeddings available directly in Google Earth Engine, designed to be used as a universal feature space for a wide variety of mapping tasks. This is distinct from Google Research's Remote Sensing Foundations (RSF) models, which are vision and vision-language transformers pre-trained on high-resolution (0.1m - 10m) imagery that can be fine-tuned or prompted on new user imagery \cite{rsfm-zeroshot,rsfm-mae,rsfm-fewshot}. The vision and vision-language RSF models are pre-trained with MAE and contrastive learning, respectively, followed by supervised learning for selected downstream tasks. Both the AlphaEarth Foundations embeddings and the Remote Sensing Foundations models are key parts of Google's Earth AI \cite{earthai}.

Table~\ref{model-comp} provides a comparative summary of all of these models, indicating a diverse landscape and a healthy and rapidly advancing field. While architectural and pre-training specifics differ, the overarching trend is a move towards leveraging vast, global, unlabeled or weakly-labeled datasets to create powerful, generalizable backbones that the applied research community can readily adapt.

\begin{sidewaystable}
\caption{Comparison of prominent Geospatial Foundation Models.}
\label{model-comp}
\centering
\scriptsize
\renewcommand{\arraystretch}{1.5} 

\begin{tabular}{>{\RaggedRight}p{2.0cm}>{\RaggedRight}p{2.0cm}>{\RaggedRight}p{4.0cm}>{\RaggedRight}p{2.25cm}>{\RaggedRight}p{2.0cm}>{\RaggedRight}p{2.0cm}>{\RaggedRight}p{2.0cm}>{\RaggedRight\arraybackslash}p{2.0cm}}

\toprule 
\textbf{Model} & \textbf{Lead \newline Institution} & \textbf{Key Focus/Innovation} & \textbf{Pre-Training \newline Strategy} & \textbf{Pre-Training \newline Data} & \textbf{Pre-Training \newline Image \newline Resolution} & \textbf{License} & \textbf{Model Size (\# of \newline Parameters)} \\
\midrule 

SatMAE & Stanford & Pioneered multi-temporal, multi-spectral pre-training with specific position encodings & Reconstruction & Sentinel-2 \newline imagery, time & 10m - 60m & CC-BY-NC 4.0 & 307M \\
\addlinespace

ScaleMAE & Berkeley & Scale-aware MAE with GSD-based position encoding and multi-scale reconstruction decoder & Reconstruction & FMoW RGB \newline imagery, GSD & 0.3 - 10m+ & CC-BY-NC 4.0 & 323M \\
\addlinespace

Cross-Scale MAE & U. Tennessee \newline Knoxville & Scale augmentation and cross scale consistency losses & Reconstruction, \newline Contrastive & FMoW RGB \newline imagery & 0.2m - 30m & CC-BY-NC 4.0 & 307M \\
\addlinespace

Prithvi-EO-2.0 & IBM/NASA & Multi-temporal model explicitly encoding time and location metadata & Reconstruction & HLS imagery, \newline location, time & 30m & MIT, \newline Apache 2.0 & 300M \& 600M \\
\addlinespace

Clay & Clay \newline Foundation & Multi-temporal, multi-spectral imagery from any sensor at any resolution & Reconstruction & General EO & Variable & Apache 2.0 & Not specified \\
\addlinespace

DINOv3 & Meta AI & Generalist vision model with strong transfer from web images; \newline Separate model pre-trained on high-resolution satellite imagery & Self-distillation & Web images or \newline Maxar RGB imagery (for satellite model) & 0.6m \newline (for satellite model) & Custom License & Up to 7B \\
\addlinespace

AlphaEarth & Google & Produces global, static "embedding field" from multi-modal, multi-temporal inputs & Multi-Loss \newline (Reconstruction, \newline Self-distillation, \newline Contrastive) & Optical, SAR, LIDAR, text, climate, time, location & 10m (output \newline embeddings) & Model is \newline Proprietary, \newline Embeddings are \newline CC-BY-4.0 & 480M \\
\addlinespace

Remote \newline Sensing \newline Foundations & Google & Fine-tunable and promptable models for high-resolution aerial imagery & Reconstruction, \newline Contrastive, \newline Supervised & High-resolution aerial imagery, text & 0.1m - 10m & Proprietary & 300M \& 400M \\

\bottomrule 

\end{tabular}
\end{sidewaystable}

\section{A Taxonomy of Geospatial Foundation Model Applications}
The different pre-training paradigms unlock distinct and transformative capabilities. This section surveys the primary downstream applications enabled by GeoFMs.

\subsection{Image Classification}
Scene-level image classification is one of the most fundamental tasks in geospatial analysis. Here, the GeoFM assigns a single class label to an image from a pre-defined, closed set of classes. A user could take a pre-trained GeoFM backbone, attach an image classification head, and fine-tune it on a labeled dataset so that the model learns how to categorize scenes. For example, a fine-tuned GeoFM could distinguish between "industrial buildings", "residential buildings" and "annual crops" at the scene level.

\subsection{Image Segmentation}
Image segmentation extends classification to the pixel level, assigning a class label to every pixel in the image from a closed set of pixel types. This allows for the precise delineation of objects and regions in the image, useful for outlining water bodies, roads, or individual buildings. A user could attach an image segmentation head to a pre-trained backbone and fine-tune it, teaching the model how to produce a pixel-wise classification map (segmentation mask). For example, a fine-tuned GeoFM could produce a segmentation mask that precisely outlines airplanes on a tarmac, distinguishing both from adjacent buildings and vegetation.

\subsection{Object Detection}
Object detection involves identifying the location of individual objects in an image and classifying what type of objects they are, based on a pre-defined, closed set of object types. A user could fine-tune an object detection head attached to a pre-trained GeoFM backbone to teach the model how to output bounding boxes for the specific objects of interest, as well as each bounding box's individual type (e.g., "pier", "warehouse").

\subsection{Open-Vocabulary Image Analysis}
Contrastively-trained vision-language models have provided a revolutionary capability, shattering the "closed-set" limitation of the traditional image analysis tasks described above. Rather than limiting the model's analysis to a closed set of image classes, pixel classes, or object types, the model can process the image based on the user's description of the classes and object types they are interested in, specified in natural language.

\begin{figure}[htbp]
\centering
Prompt: "building"\\
\includegraphics[width=0.95\textwidth]{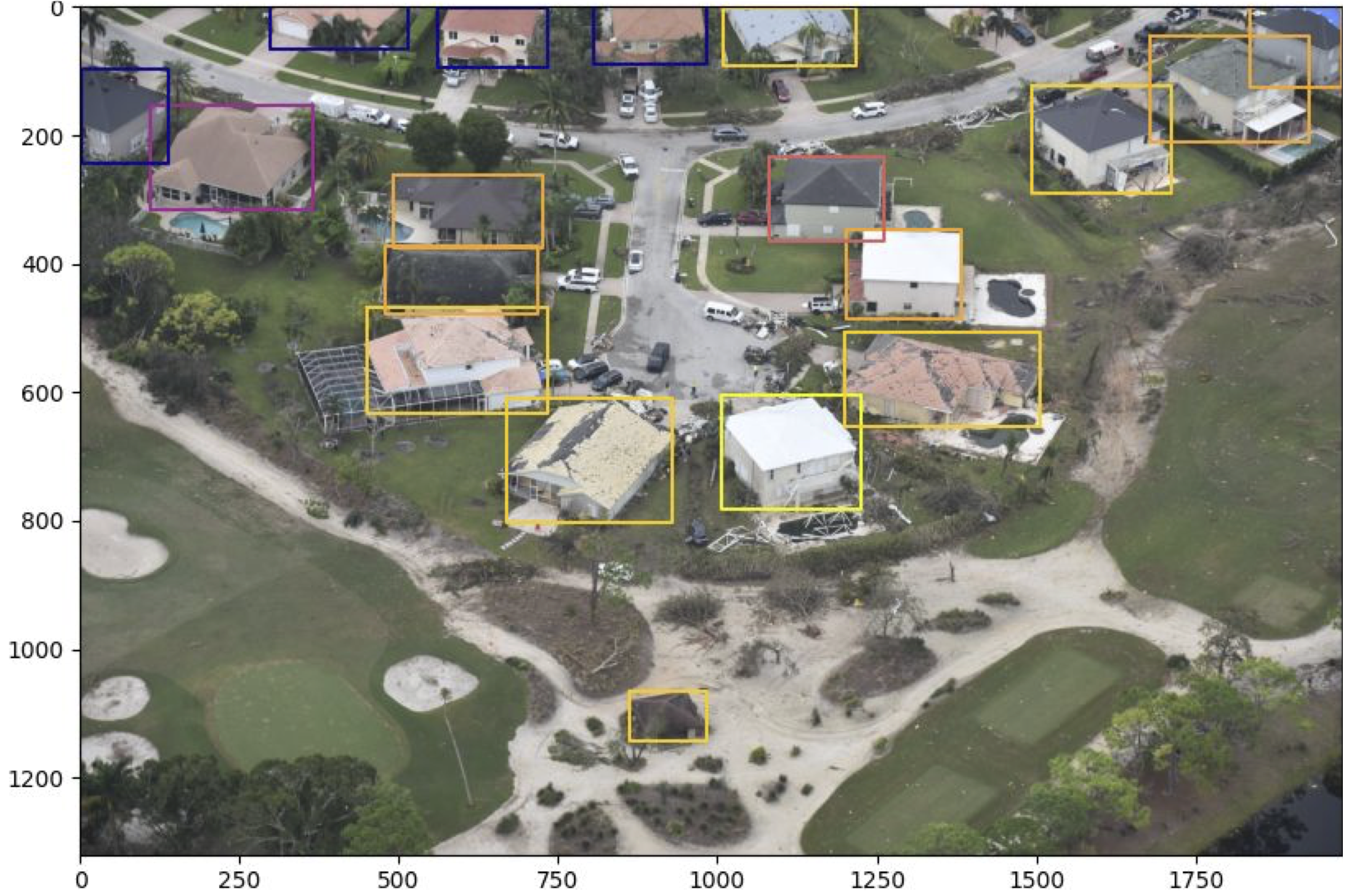}\\
Prompt: "golf course"\\
\includegraphics[width=0.95\textwidth]{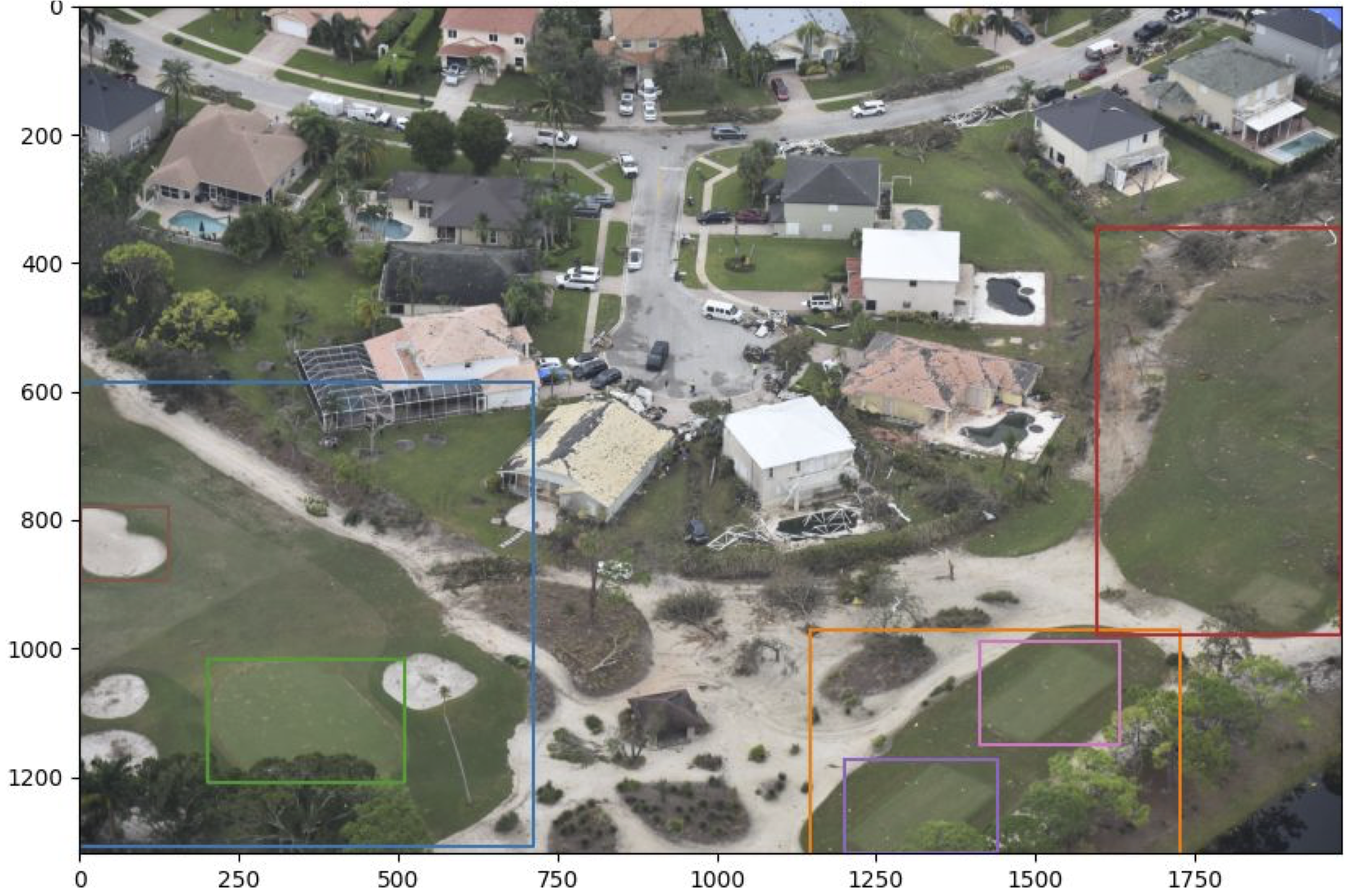}
\caption{Example of open-vocabulary object detection. The same vision-language GeoFM is prompted to detect (Top:) the text string "building" and subsequently (Bottom:) the text string "golf course", demonstrating the ability to dynamically query the image for different object types without fine-tuning. Images courtesy of Bellwether ~\cite{geospat-reas-blog}.}
\label{open_voc_obj_det}
\end{figure}

For example, in the case of open-vocabulary object detection, a user can prompt the model with a natural language text query, such as "building" or "golf course". The model can then identify and locate the corresponding objects in the image, even if it was never explicitly trained on those object types (Fig.~\ref{open_voc_obj_det}). This capability provides enormous flexibility for operational scenarios where analysts may need to search for novel or unexpected objects (e.g., "damaged roof with blue tarp") with little to no advanced preparation.

\subsection{Change Detection}
Change detection is a critical task in remote sensing. It can be used to identify differences in a geographic area over time by leveraging a GeoFM backbone to produce powerful embeddings. For example, the embedding for an image of a specific location at time $T_1$ can be compared to the embedding of an image of the same location at time $T_2$. A simple mathematical operation (e.g., cosine similarity) on the two embeddings can provide a quantitative measure of change. A large distance implies a significant change has occurred, which can be used to flag the area for further analysis by a human expert.

\subsection{Similarity Search}
The embeddings produced by GeoFMs can also be used for large-scale similarity searches across vast imagery archives. An analyst can first use the backbone to generate an embedding for a new image of interest, and then search a pre-computed database of embeddings to retrieve the most visually similar images from their archive. In addition, using the vision-language GeoFMs, the analyst's search can even be multimodal, comparing their text prompt embedding to the image embeddings in their archive. This is a powerful tool for finding other examples of rare objects or land-use types, again using simple mathematical operations (e.g., cosine similarity).

\section{Operationalizing Geospatial Foundation Models}
While the promise of GeoFMs is significant, their operationalization requires navigating a complex landscape of strategic trade-offs. Key decisions encompass the choice of deployment pattern (online inference for real-time needs versus offline batch processing of large archives) and the methodology for continuous evaluation to both validate model performance pre-deployment and to monitor model performance for degradation in production. Underpinning all of this is the degree of adaptation required to meet mission requirements. This section outlines a framework for navigating these critical operational decisions.

\subsection{A Taxonomy of Adaptation Strategies}
The decision to adapt a pre-trained GeoFM is not binary, but a selection from a spectrum of strategies, each with distinct tradeoffs in performance, cost, and labeled data requirements. We propose a taxonomy of four main adaptation strategies:
\begin{enumerate}
    \item \textbf{Zero-Shot Inference:} In this approach, the pre-trained GeoFM is used directly with no further fine-tuning. This typically involves using the backbone as a powerful feature extractor, whose embeddings can then be used for downstream tasks like change detection or similarity search. Alternatively, for vision-language GeoFMs, this could involve direct text-based prompting in natural language, as shown in Figure~\ref{open_voc_obj_det}.
    \item \textbf{Few-Shot Inference:} Once again, the frozen, pre-trained GeoFM is used with no fine-tuning to generate initial object proposals. These proposals are then refined in real-time by training a lightweight, compact classifier (e.g., support vector machine or multi-layer perception) on only a handful of actively selected user-annotated examples (the few shots). This cascading architecture/active learning approach is crucial for resolving the visual ambiguity of natural language queries ("small yacht" versus "fishing boat") to achieve a high accuracy with minimal annotation overhead \cite{rsfm-fewshot}.
    \item \textbf{Head-Only Fine-Tuning before Inference:} The pre-trained backbone remains frozen, and only a new, task-specific head is fine-tuned on a small, labeled dataset using supervised learning techniques. Inference then follows.
    \item \textbf{Full Fine-Tuning before Inference:} The entire model, including both the pre-trained backbone and the new, task-specific head, is fine-tuned on a small, labeled dataset, potentially with parameter-efficient techiques like Low-Rank Adaptation (LoRA) to reduce computational burden \cite{lora}. This transfer learning allows the backbone's deep feature representations (embeddings) to adapt more specifically to the nuances of the new dataset, in better support of the new task-specific head. The fine-tuned model can then be used for inferencing.
\end{enumerate}

\subsection{Performance-Cost Analyses}
Deciding which adaptation strategy to pursue requires a deliberate performance-cost analysis. In regards to fine-tuning, a user can perform a head-only fine-tuning or use a full fine-tuning strategy. While a full fine-tuning may yield the highest accuracy, its benefits may or may not justify its computational expense. We recommend a three-step framework to determine the optimal fine-tuning strategy for a specific use case and budget.

\begin{figure}[htbp]
\centering
\includegraphics[width=0.48\textwidth]{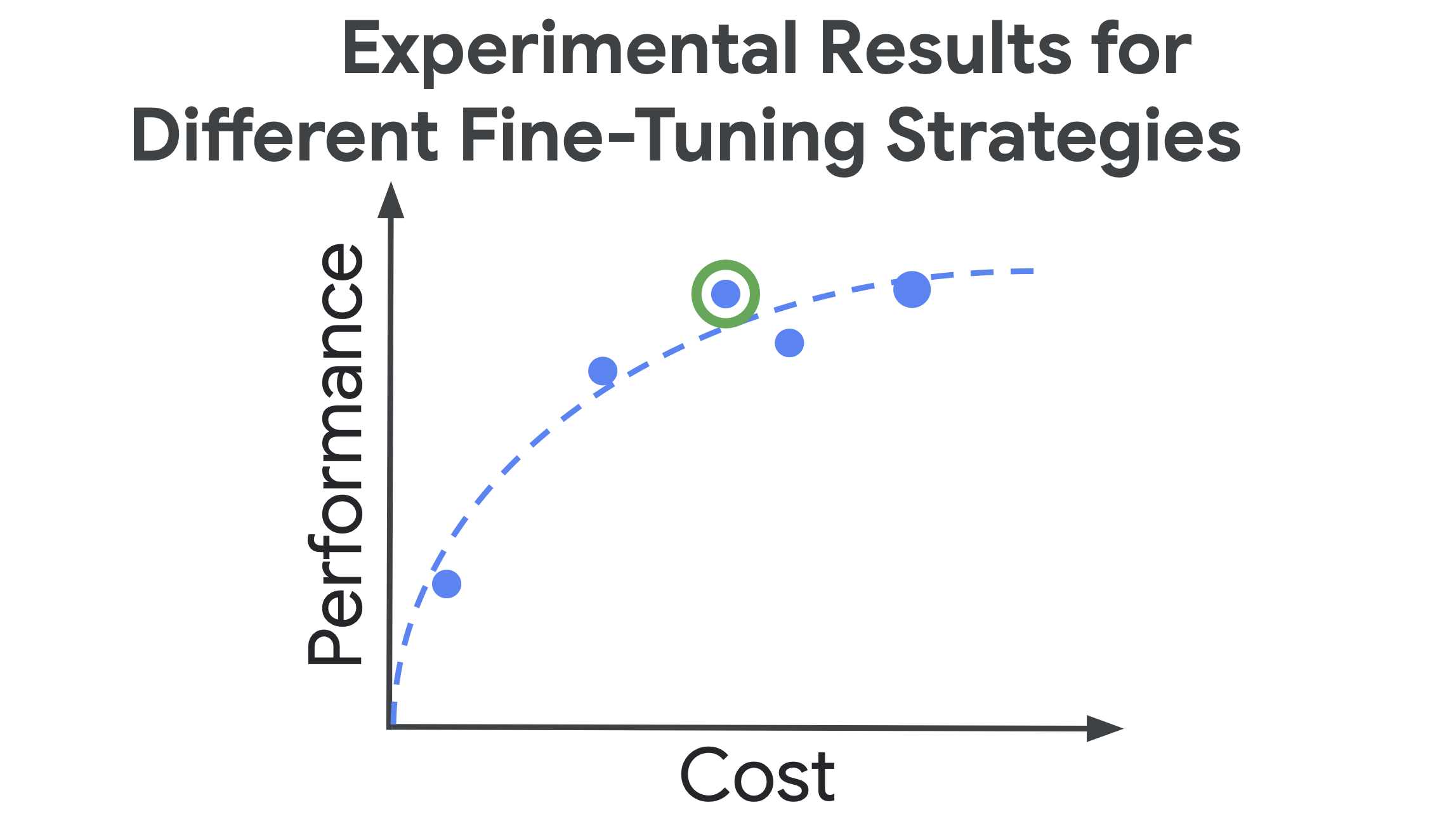}
\includegraphics[width=0.48\textwidth]{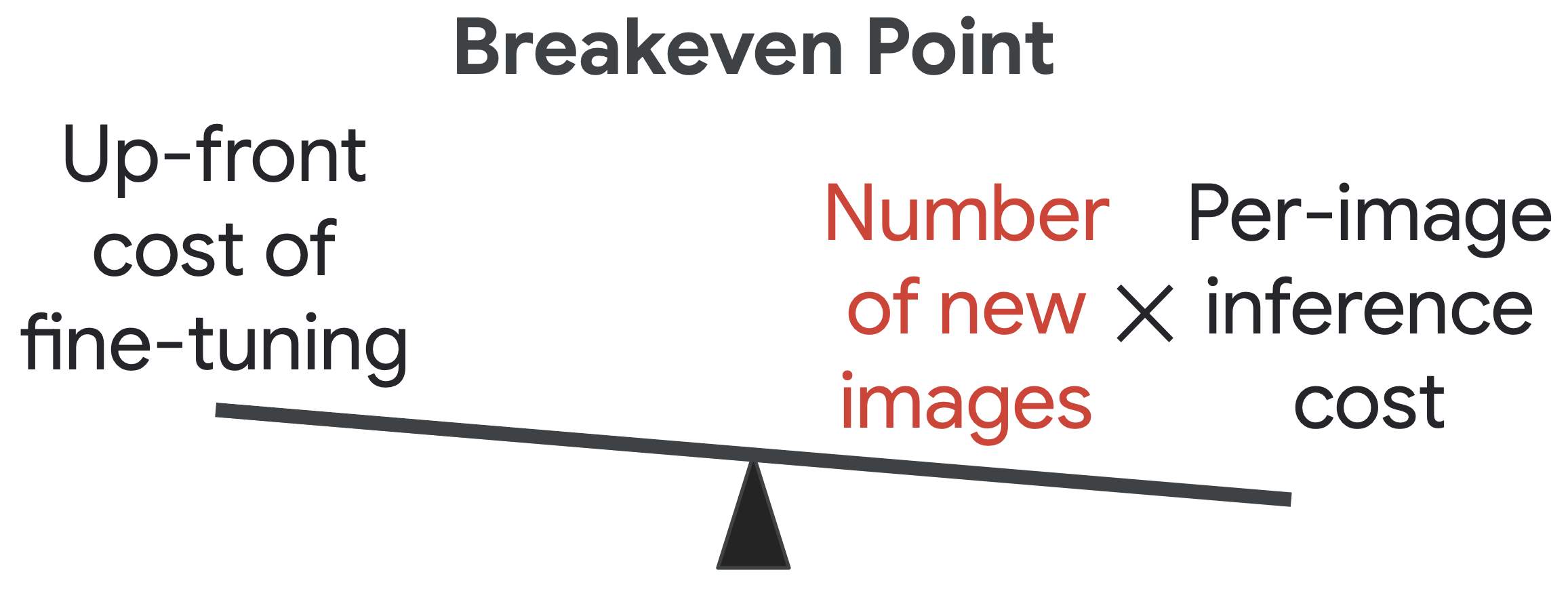}
\caption{Conceptual illustrations for performance-cost analyses. (Left:) A notional curve showing the trade-off between model performance and cost for different fine-tuning strategies. (Right:) A conceptual diagram of the breakeven point, where the up-front, fixed cost of fine-tuning is surpassed by the recurring, variable costs of inference over many images.}
\label{perf_cost}
\end{figure}

\begin{enumerate}
    \item \textbf{Empirical Benchmarking:} Organizations should conduct a series of controlled experiments to quantify the performance of each adaptation strategy on a representative validation set. This involves comparing the relevant performance metric (e.g., accuracy for image classification, mean intersection-over-union for image segmentation, F1 or mean average precision for object detection) for head-only fine-tuning and full fine-tuning strategies, \textit{while also measuring the computational resources (e.g., GPU- and CPU-hours) required for each strategy}. These experiments, ideally run in a cloud environment to explore various hardware configurations, establish an empirical performance-versus-cost curve (Fig.~\ref{perf_cost}, left).
    \item \textbf{Breakeven Point Calculation:} The analysis must account for both the \textit{up-front} cost of fine-tuning and the \textit{recurring} cost of inference. Organizations should select their preferred fine-tuning strategy from step 1 (that which produced a model with sufficiently high performance from acceptable computational resources) and then run a second series of experiments to determine the additional computational resources needed to perform inferencing on multiple images at different scales (e.g., 100 images, 1,000 images, 10,000 images). With this information, organizations can then calculate the \textit{"breakeven point", the number of images for which the cumulative cost of inferencing surpasses the up-front cost of fine-tuning the model} (Fig.~\ref{perf_cost}, right). This step establishes the return on investment for fine-tuning, which is valuable data for the next time fine-tuning is required.
    
    Furthermore, when calculating this breakeven point, organizations should also consider the up-front costs of preparing for the fine-tuning job, in addition to the costs of running the fine-tuning job itself. These additional costs would then be included in the up-front costs of Fig.~\ref{perf_cost} (right), such as the costs for collecting the training, validation, and test data; curating, labeling, and balancing those datasets; and employing or contracting for the deeper Machine Learning (ML) expertise needed for fine-tuning a transformer-based model.
    \item \textbf{Strategic Resource Allocation:} The previous two steps provide the data needed for strategic decision-making. The up-front cost of fine-tuning may align with the organization's capital expenditures (CapEx), while the recurring costs of inference may align with operational expenditures (OpEx). This data can also facilitate partnership discussions with cloud service providers, who may be willing to provide support for the infrequent fine-tuning phases in anticipation of longer-term inference workloads.
\end{enumerate}

\subsection{The Broader MLOps Ecosystem}
A GeoFM is not a complete solution, but rather a single, powerful component within a larger, human-driven operational system. Beyond the model adaptation strategy, a successful GeoFM deployment requires a multi-disciplinary team to manage the end-to-end MLOps lifecycle. Using the post-hurricane damage assessment scenario as an illustrative case, the key roles include:
\begin{itemize}
    \item \textbf{User Experience (UX) Scientists}, who elicit requirements from stakeholders. They would refine the problem definition beyond simply "finding damaged buildings" to identifying more specific, critical needs, such as "detecting impassable roads due to large-scale debris." They would also determine the operational cadence and latency requirements needed to support downstream operations.
    \item \textbf{Data Scientists}, who perform data curation and oversee its labeling. They would create the small, labeled dataset needed for fine-tuning, overseeing the annotation of images not just for "undamaged buildings", "mildly damaged buildings", and "heavily damaged buildings", but also for "road-blocking debris", all while ensuring dataset quality and balance.
    \item \textbf{ML Scientists}, who select and implement the optimal adaptation strategy. This could involve fine-tuning the GeoFM on the new, labeled dataset. For vision-language GeoFMs, this could also involve evaluating the zero-shot or few-shot performance of different prompts. They might also apply post-training optimization techniques, such as quantization, to compress the model for deployment on resource-constrained edge devices.
    \item \textbf{Deployment Engineers}, who are responsible for model serving and long-term model maintenance. Their mandate is often to deploy to specific computing environments, requiring models optimized to specific hardware. They would serve the finalized model via a scalable API and implement robust monitoring for data and concept drift, which could trigger a re-fine-tuning workflow or a re-prompting analysis, if the model's performance degrades over time.
\end{itemize}

\subsection{Limitations and Open Research Challenges}
While the GeoFM paradigm is powerful, it is not a panacea. Several significant challenges must be addressed to realize the full potential of these models.
\begin{itemize}
    \item \textbf{Data Bias and Representation:} A fundamental concern is the potential for bias in the massive pre-training datasets. Unchecked, these datasets may have significant geographical imbalances (e.g., over-representation of North American and European landscapes) or seasonal biases, which can lead to performance disparities and reduced generalization capability when the models are applied to underrepresented regions or conditions.
    \item \textbf{Computational Accessibility:} The immense computational cost of pre-training a GeoFM is a primary barrier, largely concentrating foundation model development within large industrial labs. While the "separation of duties" paradigm mitigates this, the cost of large-scale inference remains substantial, posing a hurdle for academic labs and non-profit organizations without access to large-scale GPU resources.
    \item \textbf{Principled Uncertainty Quantification:} For high-stakes applications like disaster response, a model's prediction is insufficient without a reliable estimate of its uncertainty. Developing methods that can produce calibrated, spatially-explicit uncertainty maps is critical for building user trust and enabling decision-makers to know what the model does not know.
    \item \textbf{Interpretability and Explainability:} The black-box nature of these models remains a significant barrier to adoption in critical domains. Further research is needed into methods that can explain why a model made a particular decision (e.g., classifying a detected building as mildly damaged). This information is essential for debugging, validating model reasoning against domain experts, and ensuring trustworthy deployments.
\end{itemize}

\section{The Future: Agentic Geospatial Reasoning}
The ultimate trajectory for GeoFMs lies beyond performing discrete tasks and towards enabling holistic Geospatial Reasoning \cite{earthai,geospat-reas-blog}. This vision involves using a highly capable, multi-modal LLM, such as one from the Gemini, GPT, Claude, or other LLM families, as an intelligent orchestrator that leverages a suite of tools, including GeoFMs, to solve complex problems posed in natural language.

\begin{figure}[htbp]
\centering
\includegraphics[width=\textwidth]{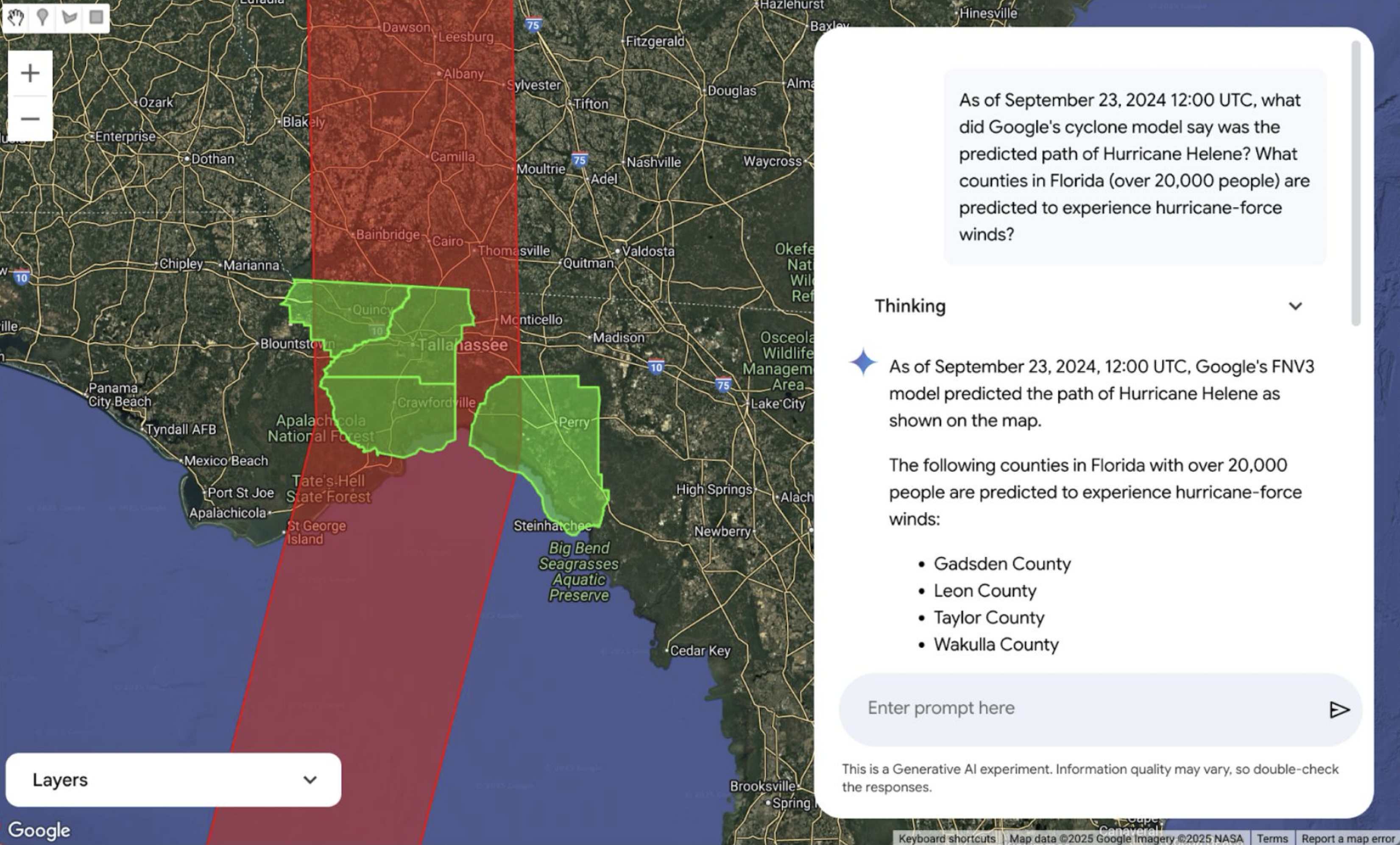}
\caption{A conceptual user interface for an agentic geospatial reasoning system. A user tasks the system via a natural language text prompt. The system then orchestrates the use of datasets and tools, including GeoFMs, to generate an analytical output. The user can then follow up with additional prompts in natural language \cite{earthai}.}
\label{geospat_reas}
\end{figure}

In this paradigm, an LLM agentic system can \textit{reason} about a high-level user request, \textit{decompose} it into a logical sequence of steps, and \textit{autonomously execute} the plan by invoking the appropriate data sources and software tools. Returning to the post-hurricane scenario, a disaster response manager could issue a simple natural language prompt: "Find me the neighborhoods most damaged by the recent hurricane." An agentic system \cite{geospat-reas-blog} could reason about this request and then define and execute a multi-step workflow:

\begin{enumerate}
    \item \textbf{Situational Awareness:} The LLM would first consult weather models and map data to identify the hurricane's landfall zone and define a geographic region of interest (Fig.~\ref{geospat_reas}).
    \item \textbf{Data Retrieval:} The LLM would then write and send a query to an imagery archive for recent, post-hurricane satellite imagery covering that specific region.
    \item \textbf{Tool Invocation:} The LLM would select the appropriate tool, such as a vision-language GeoFM, and use it to perform batch inference on the retrieved imagery, performing zero-shot object detection for "buildings".
    \item \textbf{Semantic Enrichment:} For each detected building, a subsequent model or the LLM itself would assign a damage score (none, low, medium, high damage).
    \item \textbf{Synthesis and Reporting:} Finally, the LLM would aggregate the damaged building counts by census tract, sort the tracts from most to least damaged, and provide a final, actionable summary to the user.
\end{enumerate}

Geospatial Reasoning represents a paradigm shift from straightforward image perception to complex geospatial cognition, where the analytical power of GeoFMs is made accessible and composable through natural language.

\section{Conclusion}
GeoFMs represent a fundamental change in the field of applied imagery and pattern recognition (AIPR). By separating the immense cost of pre-training from the downstream, domain-specific process of fine-tuning or prompting, they democratize access to state-of-the-art AI/ML. To advance the state-of-the-art even further, this paper issues a call to action for the AIPR community on two key fronts:

\begin{enumerate}
    \item \textbf{Develop Truly Foundational Models:} The ultimate goal is a single, universal model that is invariant to image modality (e.g., EO, IR, LiDAR, and the especially challenging physics of SAR), size, resolution, off-nadir angle, geography, season, and time of day. ML scientists must explore novel architectures and pre-training recipes to create embeddings so robust that they can be adapted for any downstream task with a simple, lightweight head that can be fine-tuned on commodity hardware. One potential path is the development of modality-agnostic encoders with learned positional priors, where dedicated "modality towers" learn to translate diverse sensor data into a common latent space before processing by a shared backbone.
    \item \textbf{Establish Governing Principles for Agentic Reasoning:} As LLMs begin to orchestrate complex analytical workflows, each automated decision can introduce bias and error. We urge human-machine teaming experts to develop governing frameworks for these systems based on a set of core principles. These principles align with the established academic framework of Meaningful Human Control, which requires systems to track human intentions and be traceable to a responsible human agent \cite{human}:
    \begin{itemize}
        \item \textbf{Verifiability:} Every output must be accompanied by a transparent, auditable log of the data sources, tools, and intermediate results used.
        \item \textbf{Graded Authority:} A framework must define which decisions can be automated versus which require a human in the loop.
        \item \textbf{Criticality Interlocks:} For critical applications, the system must be defined with “interlocks” that prevent action from being taken based on the AI's outputs until a human explicitly confirms the findings.
    \end{itemize}
\end{enumerate}

By tackling these challenges, we can collectively accelerate the transition of these powerful models from research to real-world impact. Ultimately, the vision for agentic geospatial reasoning extends beyond operational decision-making into the realm of scientific discovery. An agentic system, capable of autonomously querying data, running models, and synthesizing results, becomes a new kind of scientific instrument to test complex hypotheses about our planet at a scale previously unimaginable. By framing these systems as cognitive partners in the scientific process, we can accelerate our understanding of the earth and our impact on it.

\section*{Acknowledgements}
We thank Genady Beryozkin, Nadav Sherman, George Leifman, Tomer Shekel, and Alex Ottenwess in Google Research and Paul Rivera, Zahra Navidi, and Sam Rooney in Google Cloud for their insightful discussions and invaluable feedback. We also thank David Menasche and Kel Markert in Google Public Sector for their example workflows in geospatial reasoning with open-vocabulary object detection.

\bibliographystyle{splncs04}
\bibliography{cazares-references}    

\end{document}